%% file: main.tex
\documentclass[runningheads]{llncs}
\usepackage{graphicx}
\usepackage{tikz}
\usepackage{comment}
\usepackage{subcaption}
\usepackage{amsmath,amssymb} 
\usepackage{color}

\usepackage{hyperref}
\usepackage[super]{nth}
\usepackage{amsfonts} 
\usepackage{booktabs}
\usepackage{siunitx}
\usepackage{textcomp} 
\usepackage{caption} 
\usepackage{multirow}
\newcommand{\bftab}{\fontseries{b}\selectfont}

\usepackage[accsupp]{axessibility}  

\usepackage[utf8]{inputenc}

\begin{document}
\pagestyle{headings}
\mainmatter
\def\ECCVSubNumber{362}  

\title{CAIR: Fast and Lightweight Multi-Scale Color Attention Network for Instagram Filter Removal}

\titlerunning{CAIR: Multi-Scale Color Attention Network for Instagram Filter Removal}
%
\author{Woon-Ha Yeo\index{Yeo, Woon-Ha} \and 
Wang-Taek Oh\index{Oh, Wang-Taek} \and 
Kyung-Su Kang\index{Kang, Kyung-Su} \and 
Young-Il Kim\index{Kim, Young-Il} \and 
Han-Cheol Ryu\index{Ryu, Han-Cheol}} 
\authorrunning{W. Yeo et al.}
%
\institute{Sahmyook University, Seoul, South Korea\\
\email{canal@syuin.ac.kr, mm074111@gmail.com, unerue@me.com, qhdrmfdl123@gmail.com, hcryu@syu.ac.kr}
}
\maketitle

\begin{abstract}
Image restoration is an important and challenging task in computer vision. Reverting a filtered image to its original image is helpful in various computer vision tasks. We employ a nonlinear activation function free network (NAFNet) for a fast and lightweight model and add a color attention module that extracts useful color information for better accuracy. We propose an accurate, fast, lightweight network with multi-scale and color attention for Instagram filter removal (CAIR). Experiment results show that the proposed CAIR outperforms existing Instagram filter removal networks in fast and lightweight ways, about 11$\times$ faster and 2.4$\times$ lighter while exceeding 3.69 dB PSNR on IFFI dataset. CAIR can successfully remove the Instagram filter with high quality and restore color information in qualitative results. The source code and pretrained weights are available at \url{https://github.com/HnV-Lab/CAIR}.

\end{abstract}

\section{Introduction}
\input{article/introduction}

\section{Backgrounds}
\input{article/background}

\section{Multi-Scale Color Attention Network}
\input{article/method}

\section{Experiments}
\input{article/experiments}

\section{Conclusion}
\input{article/conclusion}

\subsubsection{Acknowledgments}
\input{article/ack}

\bibliographystyle{splncs04}
\bibliography{main}
\end{document}

%% file: article/introduction.tex
\begin{figure*}[htbp]
  \makebox[\textwidth][c]{\includegraphics[width=0.9\textwidth]{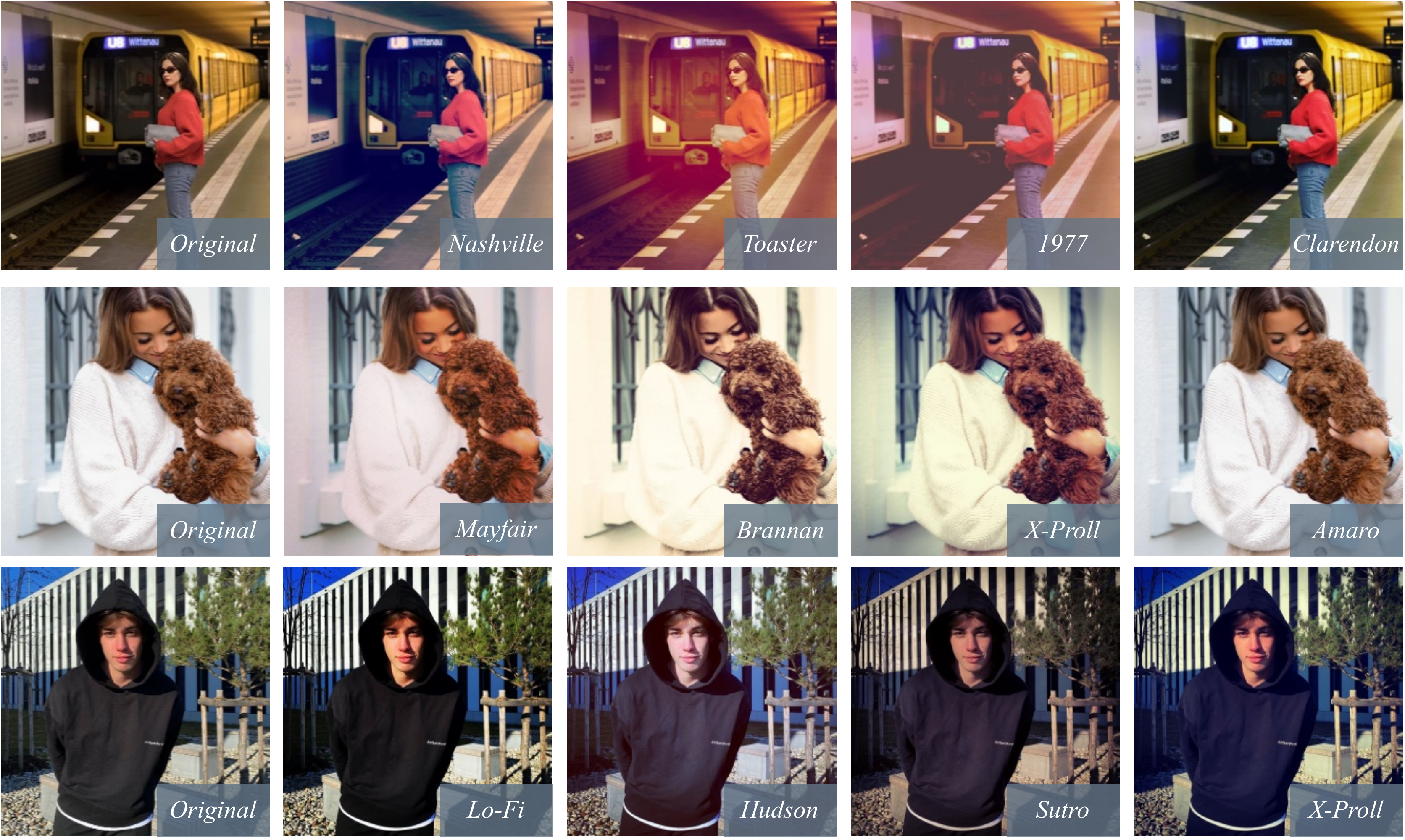}}
  \caption{Example of filtered images in Instagram Filter Fashion Image (IFFI) dataset~\cite{kinli2021instagram}. Bottom-right of each image indicates the name of the applied filter.}
  \label{fig:filter}
\end{figure*}

Photographic filters have been widely used to control the feel of photos when using analog and digital cameras. Filter as a camera accessory can be attached to the optical lens, which gives various effects to the photo (\textit{i.e.}, image), such as color conversion, color subtraction, and contrast enhancement. The popularity of smartphones has made it easier for people to take and share photos than in the days of analog and digital cameras. It has become a daily life for many people to share images on Instagram after applying various photographic effects and corrections to an image taken. These filters can digitally modify the original image (without a camera accessory) by adjusting contrast, hue, or saturation and applying blur or noise. Filters applied to images can make people feel emotional and change the photo's mood, as shown in Fig. \ref{fig:filter}. 

However, filtered images can be noise data that distracts training deep learning models from perspective of computer vision researchers. Computer vision requires a pure image state with no filters. Applying various filtering effects to raw images for data augmentation techniques can improve the generalizability of a model, but filtered images significantly degrade performance in major computer vision tasks~\cite{bianco2017artistic,hendrycks2019benchmarking} such as image classification~\cite{chen2015filter,wu2020recognizing}, detection, and segmentation~\cite{kinli2021instagram}. Therefore, filters (\textit{e.g.}, Instagram filter) need to be removed from filter images for a machine vision system to achieve non-degraded performance.

Previous studies have employed the encoder-decoder structure and adversarial training strategy to solve the filter removal problem \cite{isola2017image}. Instagram filter removal network (IFRNet) \cite{kinli2021instagram} defines the problem of removing filters from photos as a reverse style transfer \cite{huang2017arbitrary}. IFRNet consists of an encoder with VGGNet \cite{simonyan2014very} as a style extractor module and a decoder inspired by PatchGAN \cite{isola2017image}, regarding filter as additional style information. The visual effects of a filter can be directly removed by adaptively normalizing external style information in each level of the encoder \cite{kinli2021instagram}. Contrastive Instagram Filter Removal Network (CIFR)~\cite{kinli2022patch} tackles the problem by normalizing the affine parameters using VGGNet with the help of adaptive normalization. Kinli et al.~\cite{kinli2022patch} employ a multi-layer patch-wise contrastive learning strategy \cite{park2020contrastive} and propose an isolated patch sampling module that distills the content and style information \cite{kinli2022patch}. These studies showed the outstanding performance of visual understanding and removing filters; however, their structure and learning process is exceedingly complex. Thus, we desire to address the problem of filter removal efficiently with a straightforward approach.

We propose a multi-scale color attention network based on NAFNet (nonlinear activation function free network)~\cite{chen2022simple} for removing Instagram filters. The color attention modules explore different color-space between an original (non-filtered) image and a filtered image. First, we modify the architecture of NAFNet with less computational costs than the original NAFNet. Second, a color attention module inspired by~\cite{zamir2020cycleisp} is proposed to utilize the color-informative features from multi-scale input images. Third, the ensemble learning strategy is adopted to improve the performance of the results further. The proposed model is accurate, fast, and lightweight. Moreover, we consider carbon emission reduction through the lightweight model. The main contributions of this work are as follows:

\begin{enumerate}
    \item We propose an accurate, fast, and lightweight multi-scale color attention network for Instagram filter removal named CAIR, which has the remarkable qualitative and quantitative ability for Instagram filter removal.
    \item A color attention module is proposed that captures color-informative features from multi-scale input images, resulting in better filter removal performance.
    \item CAIR is a lightweight model in terms of the number of operations and parameters compared to the previous studies on filter removal.
    \item CAIR's low computational complexity improves inference speed. CAIR achieves about 11x faster runtime compared to~\cite{kinli2021instagram,kinli2022patch} without performance penalty.
\end{enumerate}

%% file: article/background.tex
\subsection{Nonlinear activation free network}
\label{sec:2.1}
\begin{figure*}[htbp]
    \centering
        \begin{subfigure}[b]{\textwidth}
        \centering
          \includegraphics[width=\textwidth]{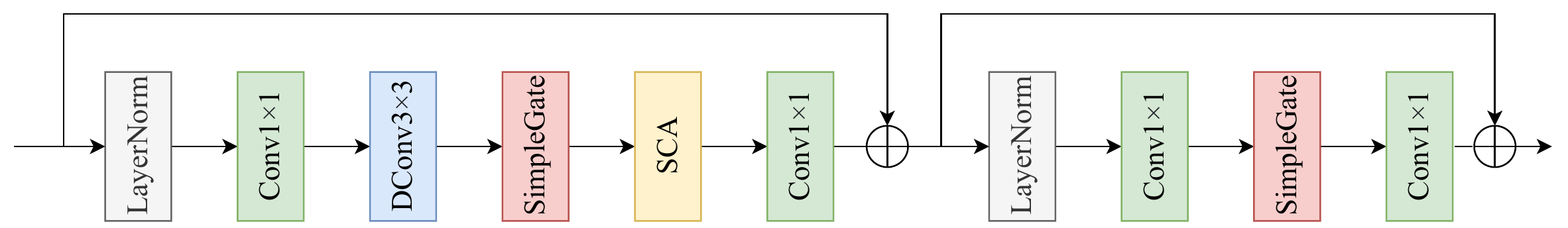}
          \caption{} 
          \label{fig:nafblock}
        \end{subfigure}
        \par\bigskip
        \centering
        \begin{subfigure}[b]{0.5\textwidth}
            \centering
            \includegraphics[width=\textwidth]{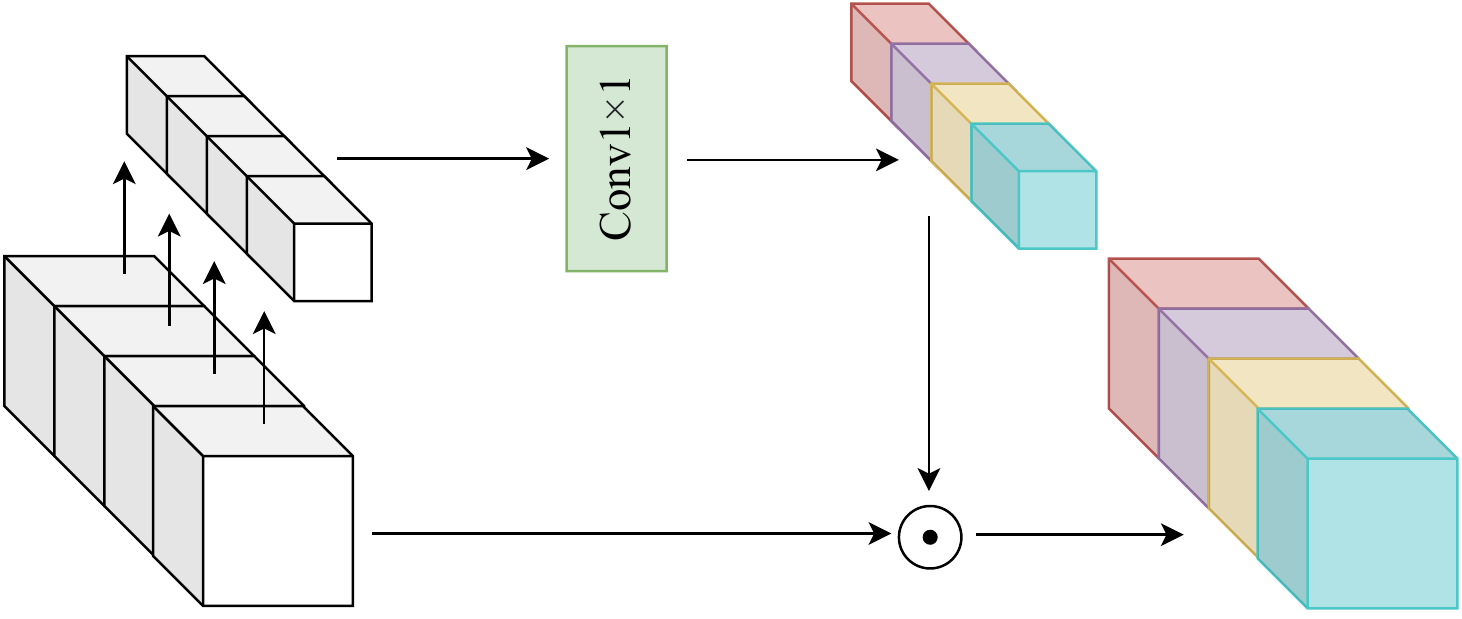}
            \caption{} 
          \label{fig:sca}
        \end{subfigure}%
        \begin{subfigure}[b]{0.5\textwidth}
        \centering
          \includegraphics[width=\textwidth]{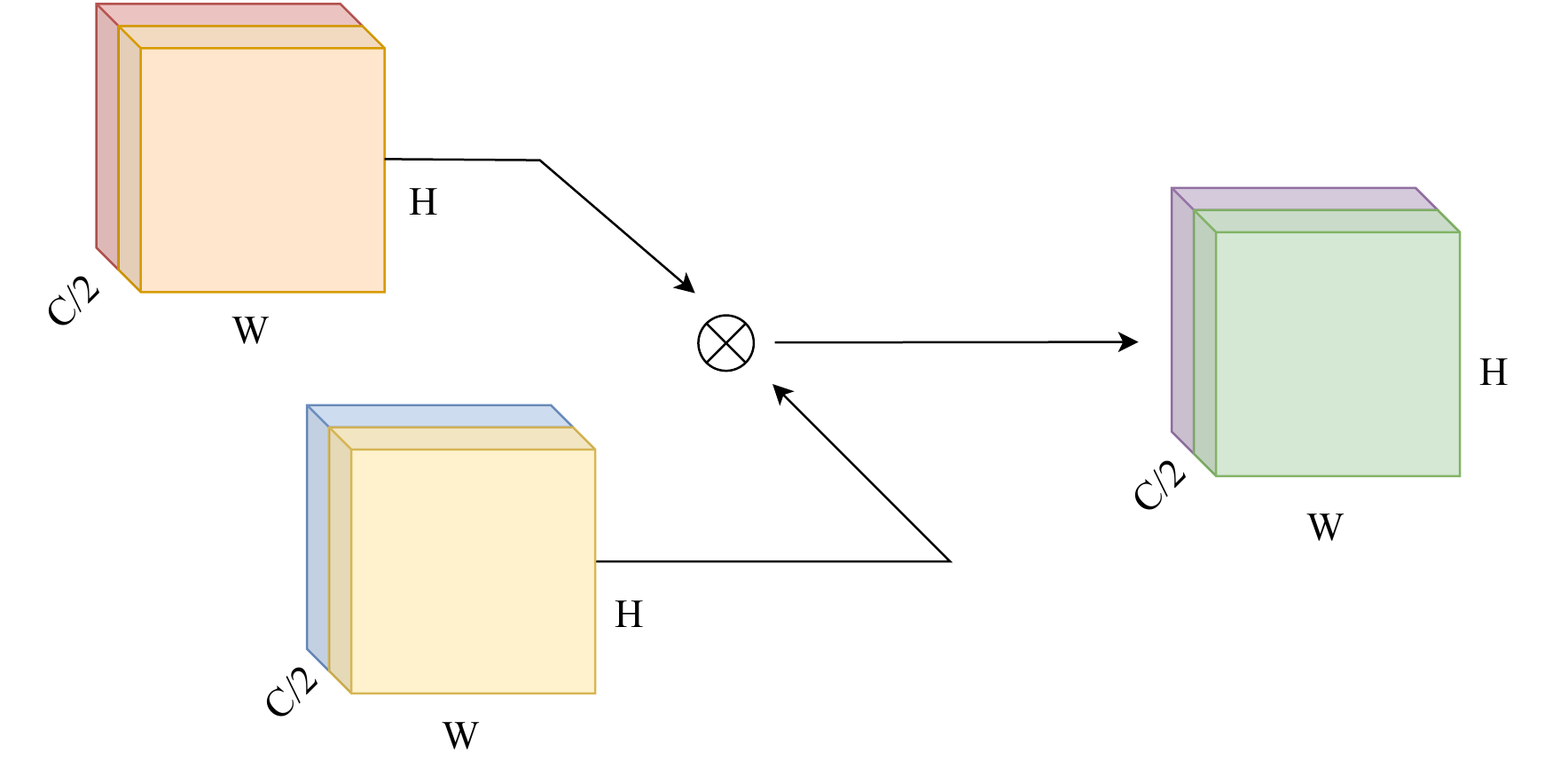}
          \caption{} 
          \label{fig:simplegate}
        \end{subfigure}
    \caption{(a) NAFBlock~\cite{chen2022simple}. (b) Simplified Channel Attention (SCA) in NAFBlock~\cite{chen2022simple}. (c) Simple Gate in NAFBlock~\cite{chen2022simple}. $\bigodot$: channel-wise multiplication, $\bigotimes$: element-wise multiplication.}
    \label{fig:nafmodules}
\end{figure*}

NAFNet~\cite{chen2022simple} is the state-of-the-art (SOTA) network in the field of image restoration (\textit{i.e.}, deblurring and denoising). Chen et al.~\cite{chen2022simple} considered two main points: 1) To enhance the performance and reduce the complexity of the network, the authors have classified inter-block and intra-block complexity. 2) They present a new direction that computer vision tasks may no longer require nonlinear activation functions. They adopt single-stage U-Net architecture~\cite{ronneberger2015u} to reduce inter-block complexity and structure simple baseline blocks for the intra-block complexity. Also, NAFNet has a critical baseline block (\textit{i.e.}, NAFBlock) that combines innovative components used in SOTA methods~\cite{tu2022maxim,wang2022uformer,zamir2022restormer}. First, layer normalization (LN)~\cite{ba2016layer} in the NAFBlock is effectively utilized for configuring the baseline block. Since LN stabilizes the training process, a large learning rate can be used during training, increasing the initial learning rate from 1e-4 to 1e-3~\cite{chen2022simple}. It also improves deblurring and denoising performance. Second, channel attention brings computational efficiency and global information to the feature maps. Third, nonlinear activation functions such as GELU~\cite{hendrycks2016gaussian} are replaced by a simple gate which is an element-wise product of feature maps. The simple gate can produce the effect of the nonlinear activation function and leads to performance gain. Finally, NAFBlock combines LN, convolution, simple gate, and simplified channel attention. NAFNet achieves SOTA performance in image restoration with its simplicity and ability. NAFBlock and its components are described in Fig.~\ref{fig:nafmodules}.

However, since it is hard to capture color information which is a crucial factor for filter removal using only channel attention, NAFNet should be further enhanced. Therefore, we propose a new NAFNet-based architecture with multi-scale color attention. In addition, we take multi-scale inputs to obtain scale-invariant features. A detailed explanation of our proposed network architecture is in Section~\ref{sec:3.1}.

\subsection{Color attention mechanism}
\label{sec:2.2}
Zamir et al.~\cite{zamir2020cycleisp} propose CycleISP to denoise sRGB images by synthesizing realistic noise data. CycleISP implements two networks: RGB2RAW network and RAW2RGB network. They convert 3-channel sRGB images to 4-channel RAW data and vice versa. In the previous study~\cite{brooks2019unprocessing}, a method for inverting the camera ISP (image signal processor) requires prior information such as color correction matrices, which degrades generalization ability. Therefore Zamir et al.~\cite{zamir2020cycleisp} propose a color correction branch and color attention unit, which provides explicit color information. The color correction branch is a convolutional neural network (CNN) where an sRGB image enters and a color-encoded deep feature is generated. he color correction branch applies Gaussian blur, 3×3 convolutional layer, two recursive residual groups (RRGs), and sigmoid activation to the sRGB image. Each RRG consists of multiple dual attention blocks, and each dual attention block contains spatial attention and channel attention modules. The color attention unit takes the output of the color correction branch and then multiplies and sums the output with the features from the previous convolutional layer in RAW2RGB. 

When constructing the color attention module in our proposed network, the RRG of the color correction branch is changed to a group of NAFBlock to increase the inference speed. The multi-scale input mentioned in Section~\ref{sec:2.1} is introduced to improve the performance. The details of the proposed color attention module are discussed in Section~\ref{sec:ca}.

\subsection{Ensemble learning}
Ensemble learning aims to achieve more synergistic effects than individual training results. In the field of MRI super-resolution, Lyu et al.~\cite{lyu2020mri} propose a GAN (generative adversarial network) architecture and train five GAN models for five differently degraded inputs. Then CNN model is used to integrate all these images. The CNN takes the outputs of five different GAN models as input to predict the final output. Ensemble learning reports the strength of the detail textures, avoids artifacts, and dramatically improves results from 0.87-0.88 to 0.95 in the structural similarity index measure (SSIM) value. This result demonstrates that using ensemble learning enhances the quality of results. Hence, we use the same ensemble learning strategy to improve further the effects of Instagram filter removal, which will be discussed in Section~\ref{sec:ensemble}.

%% file: article/method.tex
In this section, we start with the overall architecture of the proposed fast and lightweight multi-scale color attention network for Instagram filter removal (CAIR). Then we give the detailed configurations of the proposed color attention (CA) module.

\begin{figure*}[ht]
  \includegraphics[width=\textwidth]{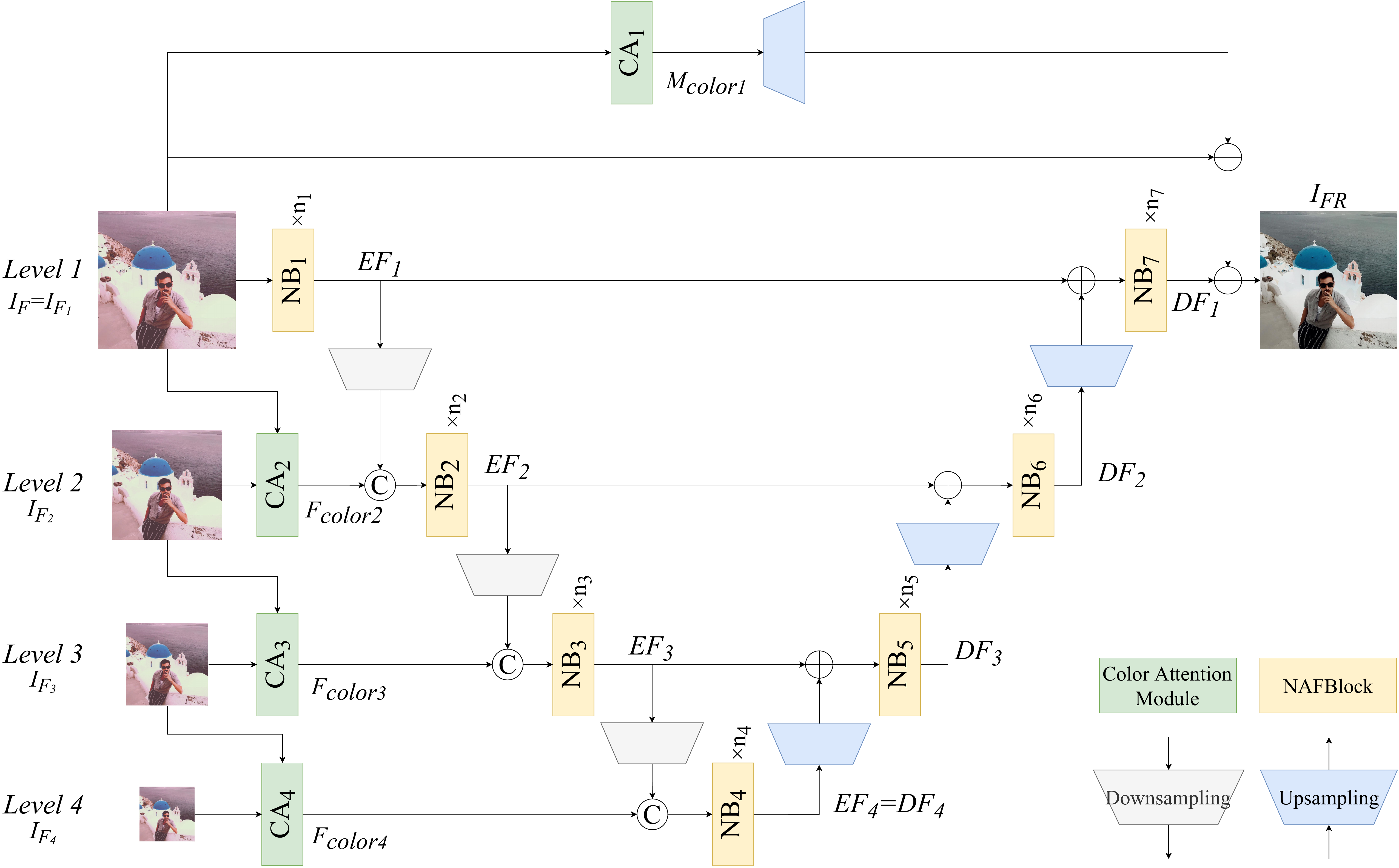}%
  \caption{The main architecture of the proposed CAIR. \textcopyright: channel-wise concatenation.}
  \label{fig:cair}
\end{figure*}

\subsection{Proposed CAIR architecture}
\label{sec:3.1}
We propose CAIR that fully utilizes multi-scale and color-attentive features extracted from an input image. Fig.~\ref{fig:cair} shows the overall architecture of CAIR. The architecture of CAIR is modified from NAFNet~\cite{chen2022simple}. We improve NAFNet in two aspects: 1) CA module inspired by color correction scheme~\cite{zamir2020cycleisp} is added, and 2) CAIR takes multi-scale input images. These two improvements make full use of color-attentive features so that the filters can be removed effectively.

There exist $l$ levels in the overall architecture ($l=4$ in this work). An input filter image $I_F \in\mathbb{R}^{H\!\times\!W\!\times\!3}$ is downscaled by a factor of 2 as going down to the lower level. The down-scaled image at level $k$ is denoted as $I_{F_k}\in\mathbb{R}^{\frac{H}{2^{k-1}}\!\times\!\frac{W}{2^{k-1}}\!\times\!3}$. These multi-scale images provide semantically robust features. First, an original-size input filter image $I_F = I_{F_1}$ passes through the first NAFBlock $NB_1(\cdot)$, then the encoded feature $EF_1$ is extracted
\begin{equation}
\text{\emph{EF}}_1 = \text{\emph{NB}}_1(I_{F_1}).
\end{equation}

Then given the lower-level input $I_{F_k}$ and the upper-level input ${I_{F_{k-1}}}$, a CA module denoted as $\mathit{CA}(\cdot)$ is used to extract color-attentive features from $I_{F_k}$ and $I_{F_{k-1}}$
\begin{equation}
\text{\emph{F}}_{\text{color}_k} = \text{\emph{CA}}_k(I_{F_k}, I_{F_{k-1}}),
\end{equation}
where $F_{\text{color}_k}$ is color-attentive features, and $\mathit{CA}_k$ denotes CA module at level $k$ (for $k\geq2$). $\mathit{CA}_1$ is a special case that takes only $I_{F_1}$ and outputs color attention map $M_{\text{color}_1}$ (see Eq.~\ref{eq:9}). The CA module aims to capture the color information of the input image. The details of the CA module are explained in Section~\ref{sec:ca}.

The encoded features from upper-level $\text{\emph{EF}}_{k-1}$ are downsampled (denoted as $\downarrow$), and the color-attentive features $F_{\text{color}_k}$ are concatenated. Then the concatenated features are passed to the $k^{th}$ NAFBlock $NB_k$ to extract the encoded feature at level $k$
\begin{equation}
\text{\emph{EF}}_k=\text{\emph{NB}}_k(\text{concatenate}(\downarrow \text{\emph{EF}}_{k-1}, F_{\text{color}_k})),\quad k=2,\ldots,l,
\end{equation}

\noindent where $EF_k$ is encoded features at the $k^{th}$ level. At the $l^{th}$ level, the encoded features $\text{\emph{EF}}_l$ are same with the decoded features $\text{\emph{DF}}_l$:
\begin{equation}
\text{\emph{EF}}_l=\text{\emph{DF}}_l.
\end{equation}

The decoded features for each level $k$ can be represented as
\begin{equation}
\text{\emph{DF}}_k=\text{\emph{NB}}_{8-k}(EF_k + \uparrow DF_{k+1}),\quad k=1,2,\ldots,{l-1},
\end{equation}
\noindent where $\text{\emph{NB}}_{8-k}$ indicates $(8-k)^{th}$ NAFBlock. The $(k+1)^{th}$ level decoded features $DF_{k+1}$ is upsampled (denoted as $\uparrow$) and element-wisely added with the encoded features $EB_k$ obtained at level $k$ and then passed to the $(8-k)^{th}$ NAFBlock. The sub-pixel convolution~\cite{shi2016real} is employed as the upsampling module, which converts the scale sampling with a given magnification factor by pixel translation.

Finally, the filter removed image $I_{FR}\in \mathbb{R}^{H\!\times\!W\!\times\!3}$ can be obtained by adding the last decoded features $DF_1$ and features from the two global skip connections; one is input filtered image $I_F$, and the other is color attention map $F_{\text{color}_1}$ of the input image. These global skip connections can stabilize the training of the proposed deep network and provide information-rich features to the final output
\begin{equation}
I_{FR} = I_{F} + DF_1 + \uparrow F_{\text{color}_1}.
\end{equation}

Given a training set $\{I^i_F, I^i_O\}_{i=1}^N$, which contains $N$ filtered image $I_F^i$ and their original image $I_O^i$, the goal is to minimize the loss function. Since we modified the architecture from NAFNet, the peak-signal-to-noise (PSNR) loss is selected as in~\cite{chen2022simple} for the fair comparison
\begin{equation}
L(\Theta) = -\frac{1}{N} \sum^N_{i=1} \mathit{PSNR}(H_{\mathit{CAIR}}(I_F^i), I_O^i) = -\frac{1}{N} \sum^N_{i=1} \mathit{PSNR}(I_{FR}^i, I_O^i),
\end{equation}
where $H_{\text{\emph{CAIR}}}$ and $\Theta$ denote the function of the proposed CAIR and the training parameter of the CAIR, respectively.

\begin{equation}
    \mathit{PSNR}(X, Y) = 10\cdot \log_{10} \frac{255^2}{\mathit{MSE}(X,Y)},
\end{equation}
\begin{equation}
\mathit{MSE}(X, Y) = \frac{1}{WH} \sum_{w=1}^W \sum_{h=1}^H (X_{w,h}-Y_{w,h})^2,
\end{equation}
\noindent where $W$ and $H$ are the width and height of images $X$ and $Y$.

\subsection{Color attention module}
\label{sec:ca}
\begin{figure*}[htbp]
  \centering
  \includegraphics[width=0.9\textwidth]{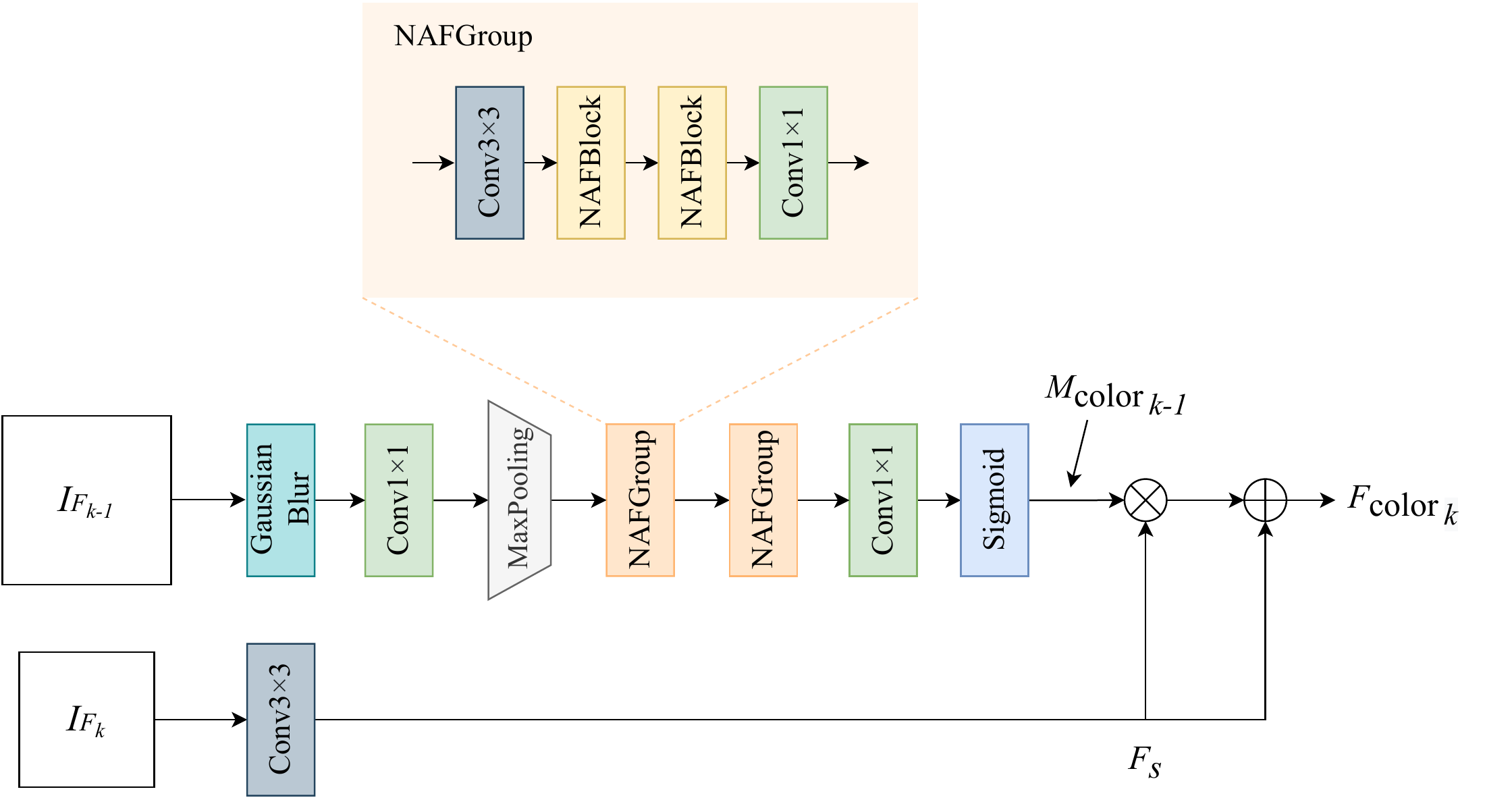}%
  \caption{The proposed color attention (CA) module. {$\bigotimes$: element-wise multiplication, $\bigoplus$: element-wise summation.}}
  \label{fig:ca}
\end{figure*}

Inspired by the color attention unit and color correction branch in~\cite{zamir2020cycleisp}, we propose a CA module for image filter removal. RRG~\cite{zamir2020cycleisp} in the color correction branch contains channel attention and spatial attention. The spatial attention can be modified to channel attention~\cite{zamir2022restormer}, and the channel attention in NAFBlock is even simplified to achieve less computational cost but better performance. Therefore, we replace the RRG with a group of NAFBlock and replace $3\!\times\!3$ convolution in the color correction branch with $1\!\times\!1$ convolution to reduce the complexity of the proposed network. In addition, our proposed color attention module takes two multi-scale input filter images and extracts color-attentive features. Our proposed CA module can learn to capture color information of multi-scale images so that it helps to get rid of filter from input filtered images. The structure of the CA module is shown in Fig.~\ref{fig:ca}.

Given $k^{th}$ level input image $I_{F_k}\in\mathbb{R}^{\frac{H}{2^{k-1}}\!\times\!\frac{W}{2^{k-1}}\!\times\!3}$ and the upper-level ($k\!-\!1^{th}$) input image $I_{F_{k-1}}$, we first extract color maps $M_{\text{color}_{k-1}}$ from the upper-level input image
\begin{equation}
\label{eq:9}
M_{\text{color}_{k-1}} = \sigma(\text{conv2}(NG_2(NG_1(\text{maxpool}(\text{conv1}(K\ast I_{F_{k-1}})))))),
\end{equation}
where $\ast$ denotes convolution operation, and $K$ is the Gaussian kernel with a standard deviation of 12~\cite{zamir2020cycleisp}. This strong blurring operation ensures that only the color information can be extracted from the input image $I_{F_{k-1}}$ whereas the structural content and fine texture come from the lower-level image $I_{F_k}$. 
After blurring, $1\!\times\!1$ convolution is applied $\text{conv1}$ followed by a max pooling (maxpool) operation. These can reduce the number of operations afterward.
$NG_1$, $NG_2$ denotes group of NAFBlocks (NAFGroup) that consists of a $3\!\times\!3$ and a $1\!\times\!1$ convolution layer, and two NAFBlocks. Another $1\!\times\!1$ convolution denoted as conv2 and a gating mechanism with sigmoid activation $\sigma$ are applied.

Once the color map $M_{\text{color}_{k-1}}$ is extracted, the structural features (denoted as $F_s$) of the lower-level image obtained by $3\!\times\!3$ convolution (denoted as conv3) are weighted through element-wise multiplication ($\otimes$) with the color map. Then element-wise summation of the structural features $F_s$ is applied to the weighted color features
\begin{equation}
    F_s = \text{conv3}(I_{F_{k}}),
\end{equation}
\begin{equation}
F_{\text{color}_k} = (M_{\text{color}_{k-1}} \otimes F_s) \oplus F_s.
\end{equation}

%% file: article/experiments.tex
\subsection{Datasets}
The AIM 2022 challenge on Instagram filter removal provided the IFFI dataset, which contains 16,000 training images, 3,000 validation images, and 2,200 test images. Training images consist of 8,000 $1080\!\times\!1080$ high-resolution (HR) images and 8,000 $256\!\times\!256$ low-resolution (LR) images. The training images include an original (unfiltered) image and images obtained by transforming the original with 15 different Instagram filters (a total of 16 versions, including the original). The validation and test images have a similar composition as training images that are divided into high and low-resolution images, and these images are comprised of 15 and 11 filtered images, respectively, without the original version. The number of Instagram filters for training and validation datasets is 15 (\textit{1977, Amaro, Brennan, Clarendon, Gingham, He-Fe, Hudson, Lo-Fi, Mayfair, Nashville, Perpetua, Sutro, Toaster, Valencia, and X-proll}), and for testing dataset is 11 (\textit{Amaro, Clarendon, Gingham, He-Fe, Hudson, Lo-Fi, Mayfair, Nashville, Perpetua, Valencia, and X-ProII}).

Additionally, we resized HR images $1080\!\times\!1080$ into $256\!\times\!256$ by resampling using pixel area relation (\textit{i.e., OpenCV} INTER\_AREA interpolation) due to image quality. 
As a result, we used three sets of images for training, including HR images, resized HR images, and LR images.
Finally, we took 24,000 images to train the proposed network CAIR and 2,200 images to test.

\subsection{Implementation details}
In the training process, data augmentation was applied to IFFI datasets~\cite{kinli2021instagram} using random flip and rotation with the probability of 0.5. In the testing process, we used the self-ensemble method in~\cite{lim2017enhanced} as a test time augmentation (TTA). The number of NAFBlocks ($n_1$-$n_7$) followed the default setting of NAFNet $(2, 2, 4, 22, 2, 2, 2)$, and the width in all NAFBlocks and convolutional layers was set to 32. The ``Downsampling'' module (represented as a gray trapezoid block in Fig.~\ref{fig:cair}) contains a convolutional layer with stride 2, and the ``Upsampling'' module (represented as a light-blue upside-down trapezoid block in Fig.~\ref{fig:cair}) is the sub-pixel convolution layer~\cite{shi2016real}. The mini-batch size was 64, and the patch size was $256\!\times\!256$ (HR images are randomly cropped to $256\!\times\!256$). We used TLSC~\cite{chu2021revisiting} to solve the problem that training patch by patch and testing with the whole image causes performance degradation~\cite{chu2021revisiting}. We used AdamW \cite{loshchilov2017decoupled} optimizer with $\beta_1=0.9, \beta_2=0.9$, weight decay $1e^{-4}$. The initial learning rate was set to $1e^{-3}$ and gradually reduced to $1e^{-6}$ with the cosine annealing schedule~\cite{loshchilov2016sgdr}. Our model was trained for up to 200K iterations. We implemented the code using PyTorch deep learning framework~\cite{NEURIPS2019_9015}. All models were conducted by four NVIDIA Tesla V100 GPUs ($4\!\times\!32$GB). We reported the number of parameters and operations, inference time on GPU for fairness, and the PSNR and SSIM value in our environment.

\subsection{Compared methods}
We compared the proposed method CAIR with SOTA Instagram filter removal models~\cite{kinli2021instagram,kinli2022patch} to show the performance of Instagram filter removal. Furthermore, to figure out the effects of the color attention module and the multi-scale input, we compared the original NAFNet, NAFNet with single input and color attention module (denoted as CAIR-S), and modified CAIR-S to take multi-scale input images (denoted as CAIR-M shown in Fig.~\ref{fig:cair}). Then we used the self-ensemble strategy~\cite{lim2017enhanced} to improve our CAIR-S and CAIR-M results further, and noted the self-ensemble applied models as CAIR-S+ and CAIR-M+. Besides, both the self-ensemble and ensemble learning strategy applied to CAIR is denoted as CAIR*. The ensemble learning is described in Section~\ref{sec:ensemble}.

\subsection{Ensemble learning}
\label{sec:ensemble}
\begin{figure*}[htbp]
  \makebox[\textwidth][c]{\includegraphics[width=0.5\textwidth]{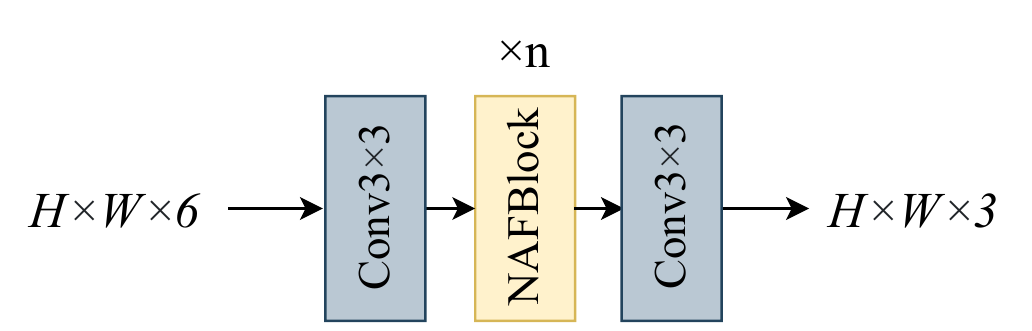}}%
  \caption{The network architecture for ensemble learning.}
  \label{fig:ens}
\end{figure*}

We used an ensemble learning strategy using CNN as in~\cite{lyu2020mri}. We propose an ensemble network composed of $3\!\times\!3$ convolution layers and $n$ NAFBlocks (in our experiment, we chose $n=3$). Fig.~\ref{fig:ens} shows the ensemble network. First, we trained CAIR-S and CAIR-M and obtained the predicted train images from the two models. After that, two output images of each model were concatenated to a single image as six channels and fed into the ensemble network. The training set was the same as the CAIR training configuration except for batch size, which is 16 here. In the testing phase, test images were predicted by two pretrained models (CAIR-S and CAIR-M), and self-ensemble was also applied. The images from the two models were concatenated and fed into the pretrained ensemble network.

\subsection{Results}

\subsubsection{Qualitative results} 
\begin{figure}[htbp]
        \centering
        \begin{subfigure}[b]{0.158\textwidth}
                \includegraphics[width=\textwidth]{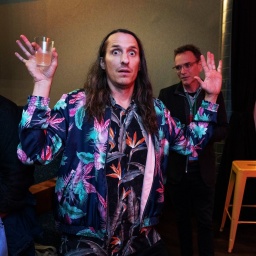}
                \includegraphics[width=\textwidth]{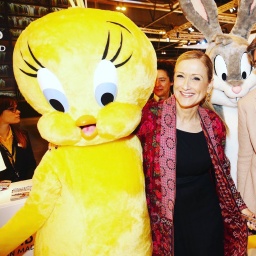}
                \includegraphics[width=\textwidth]{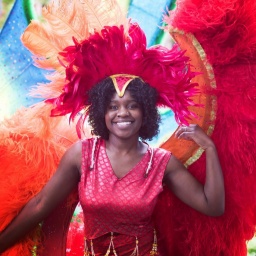}
                \includegraphics[width=\textwidth]{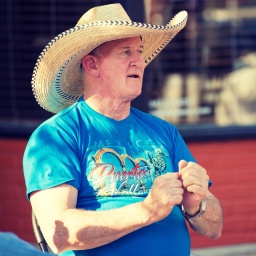}
                \includegraphics[width=\textwidth]{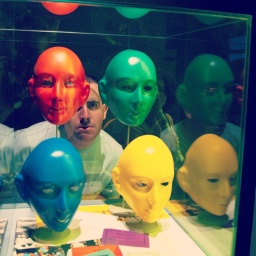}
                \label{fig:Filtered}
                \vspace*{-3mm}
                \caption{Filtered}
        \end{subfigure}  
        \begin{subfigure}[b]{0.158\textwidth}
                \includegraphics[width=\textwidth]{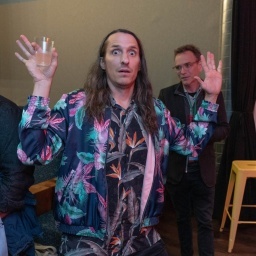}
                \includegraphics[width=\textwidth]{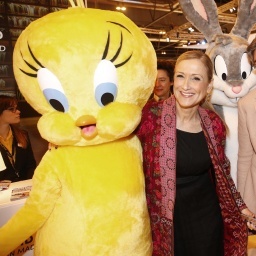}
                \includegraphics[width=\textwidth]{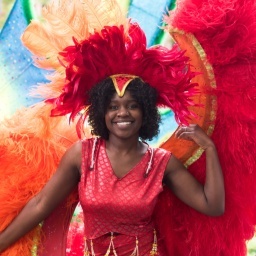}
                \includegraphics[width=\textwidth]{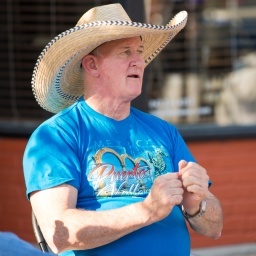}
                \includegraphics[width=\textwidth]{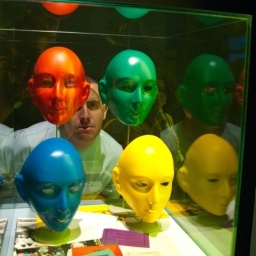}
                \label{fig:Original}
                \vspace*{-3mm}
                \caption{Original}
        \end{subfigure}       
        \begin{subfigure}[b]{0.158\textwidth}
               \includegraphics[width=\textwidth]{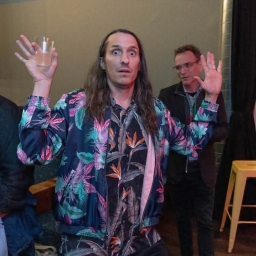}
                \includegraphics[width=\textwidth]{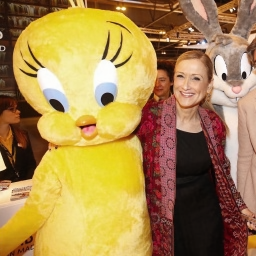}
                \includegraphics[width=\textwidth]{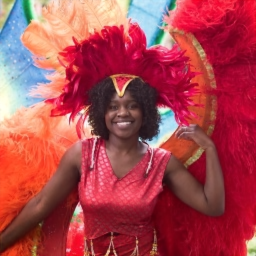}
                \includegraphics[width=\textwidth]{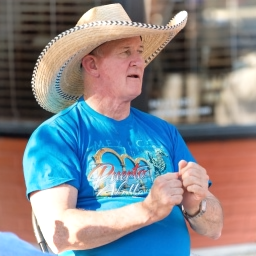}
                \includegraphics[width=\textwidth]{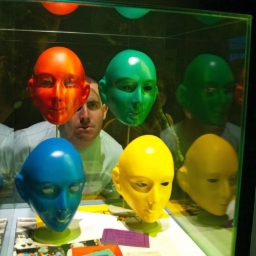}
                \label{fig:CAIR*}
                \vspace*{-3mm}
                \caption{CAIR*}
        \end{subfigure}
        \begin{subfigure}[b]{0.158\textwidth}
                \includegraphics[width=\textwidth]{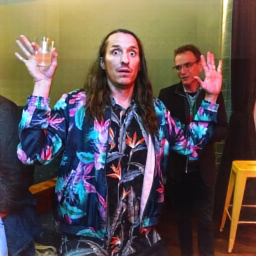}
                \includegraphics[width=\textwidth]{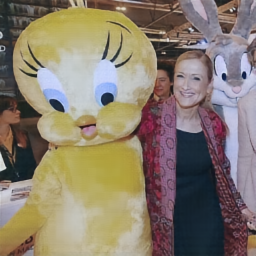}
                \includegraphics[width=\textwidth]{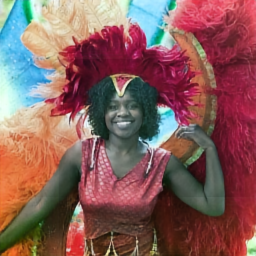}
                \includegraphics[width=\textwidth]{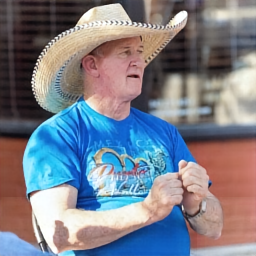}
                \includegraphics[width=\textwidth]{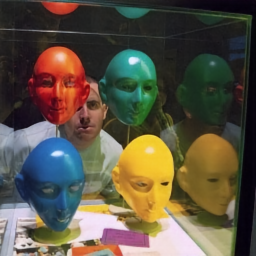}
                \label{fig:IFRNet}
                \vspace*{-3mm}
                \caption{IFRNet~\cite{kinli2021instagram}}
        \end{subfigure}
        \begin{subfigure}[b]{0.158\textwidth}
                \includegraphics[width=\textwidth]{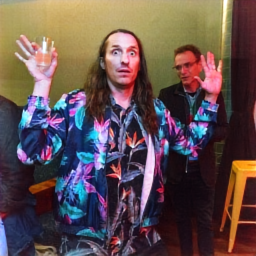}
                \includegraphics[width=\textwidth]{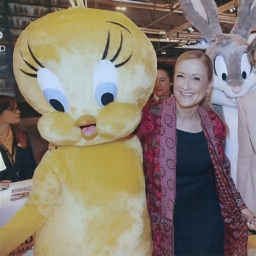}
                \includegraphics[width=\textwidth]{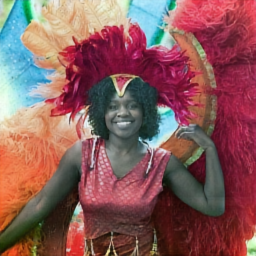}
                \includegraphics[width=\textwidth]{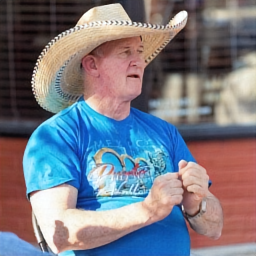}
                \includegraphics[width=\textwidth]{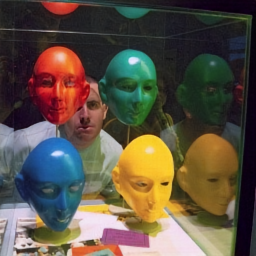}
                \label{fig:CIFR}
                \vspace*{-3mm}
                \caption{CIFR~\cite{kinli2022patch}}
        \end{subfigure}
        \begin{subfigure}[b]{0.158\textwidth}
                \includegraphics[width=\textwidth]{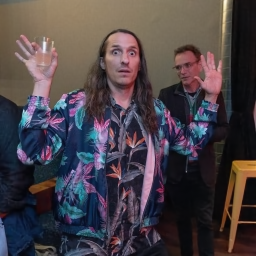}
                \includegraphics[width=\textwidth]{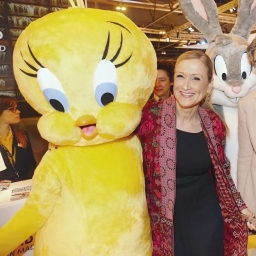}
                \includegraphics[width=\textwidth]{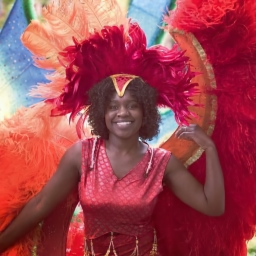}
                \includegraphics[width=\textwidth]{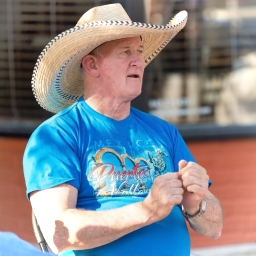}
                \includegraphics[width=\textwidth]{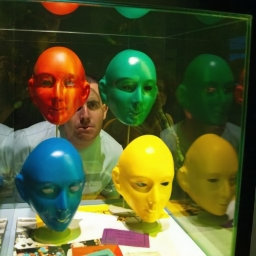}
                \label{fig:NAFNet}
                \vspace*{-3mm}
                \caption{NAFNet~\cite{chen2022simple}}
        \end{subfigure}
        \vspace*{-1mm}
        \caption{Comparison of the qualitative results on Instagram filter removal on IFFI test dataset~\cite{kinli2021instagram}. Filters applied (top to bottom): \textit{Lo-Fi}, \textit{Clarendon}, \textit{Hudson}, \textit{X-Proll}, \textit{Nashville}.}
        \label{fig:result} 
\end{figure}

The effects of the proposed color attention module can be seen in Fig.~\ref{fig:result}. It is noticeable that the skin color in the test images was restored only in CAIR* (see first, second, and third rows). Compared to the other methods, CAIR, which has the strength of the color attention module, successfully restored color information, especially chromatic colors (\textit{e.g.}, red, yellow, or green) visually. On the contrary, IFRNet and CIFR captured detailed textures rather than the proposed CAIR. In the \nth{4} row in Fig~\ref{fig:result}, the texture of the iron fence in the background was distorted on CAIR and NAFNet results. However, overall images predicted with CAIR* showed visually more satisfying results than the others.

\captionsetup[subfigure]{font=small, labelformat=empty}

\subsubsection{Quantitative results} We summarized PSNR and SSIM results, inference time, and the number of parameters and operations of several compared methods, including our CAIR, IFRNet~\cite{kinli2021instagram}, and CIFR~\cite{kinli2022patch} on the IFFI dataset~\cite{kinli2021instagram} in Table~\ref{tab:res}.
We made several observations based on the results. CAIR-S and CAIR-M outperform IFRNet and CIFR in all metrics by large margins, which validates the efficiency of the proposed method. CAIR-M also shows better PSNR and SSIM than NAFNet with fewer operations (GMACs).
In addition, the result indicates that using the self-ensemble strategy~\cite{lim2017enhanced} improved 0.24dB PSNR and about 0.002 SSIM. CAIR* achieved the best PSNR and SSIM among compared methods, meaning that the ensemble learning creates a synergy effect when combining two images predicted from CAIR-S and CAIR-M.

\begin{table}[htbp]
\begin{center}
\caption{Quantitative results on IFFI dataset~\cite{kinli2021instagram}. Obtained the available results of PSNR and SSIM from~\cite{kinli2021instagram,kinli2022patch}. Best results highlighted in bold. `+': self-ensemble~\cite{lim2017enhanced}, `*': ensemble learning strategy applied, `Runtime': inference time per image on GPU, `MACs': multiply accumulate operations per second, `Params': number of parameters.}

\setlength{\tabcolsep}{6pt}
\begin{tabular}{lccccc} \toprule
\label{tab:res}
    \text{Method} & \text{PSNR}  & \text{SSIM} & \text{Runtime (ms)} & \text{MACs (G)} &\text{Params (M)} \\\midrule
    \text{IFRNet~\cite{kinli2021instagram}} & 30.46 & 0.864 & 427.35 & 36.67 & 24.52\\ 
    \text{CIFR~\cite{kinli2022patch}}   & 29.24 & 0.888 & 420.17 & 36.67 & 24.52  \\ 
    \text{NAFNet~\cite{chen2022simple}} & 31.76 & 0.958 & 29.64 & 16.21 & 29.16\\ \midrule
    \text{CAIR-S} & 33.63 & 0.969 & 36.75 & 17.44 & 29.23\\
    \text{CAIR-S+} & 33.87 & 0.970 & 145.77 & 139.56 & 29.23\\
    \text{CAIR-M} & 34.15 & 0.969 & 45.15 & 15.23 & 13.13 \\
    \text{CAIR-M+} & 34.39 & 0.971 & 186.22 & 121.83 & 13.13\\
    \text{CAIR*} & \bftab{34.42} & \bftab{0.972}  & 25.87 & 12.4 & 0.028\\ \bottomrule
\end{tabular}
\end{center}
\end{table}

\subsubsection{Model complexity} 
\begin{figure}[htbp]
    \centering
    \begin{subfigure}[b]{0.49\textwidth}
        \includegraphics[width=\textwidth]{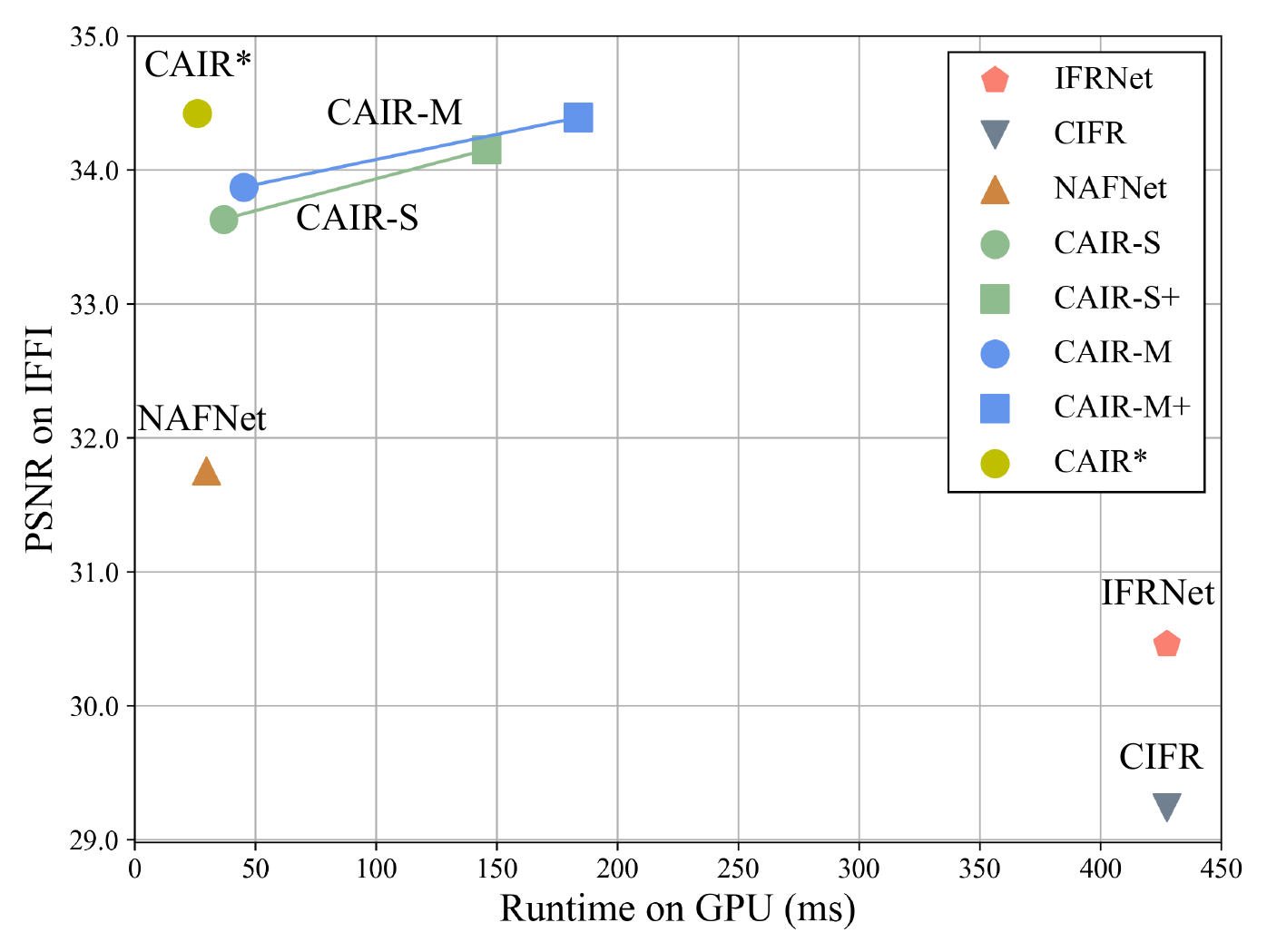}
        \subcaption{(a)}
    \end{subfigure}
    \begin{subfigure}[b]{0.49\textwidth}
        \includegraphics[width=\textwidth]{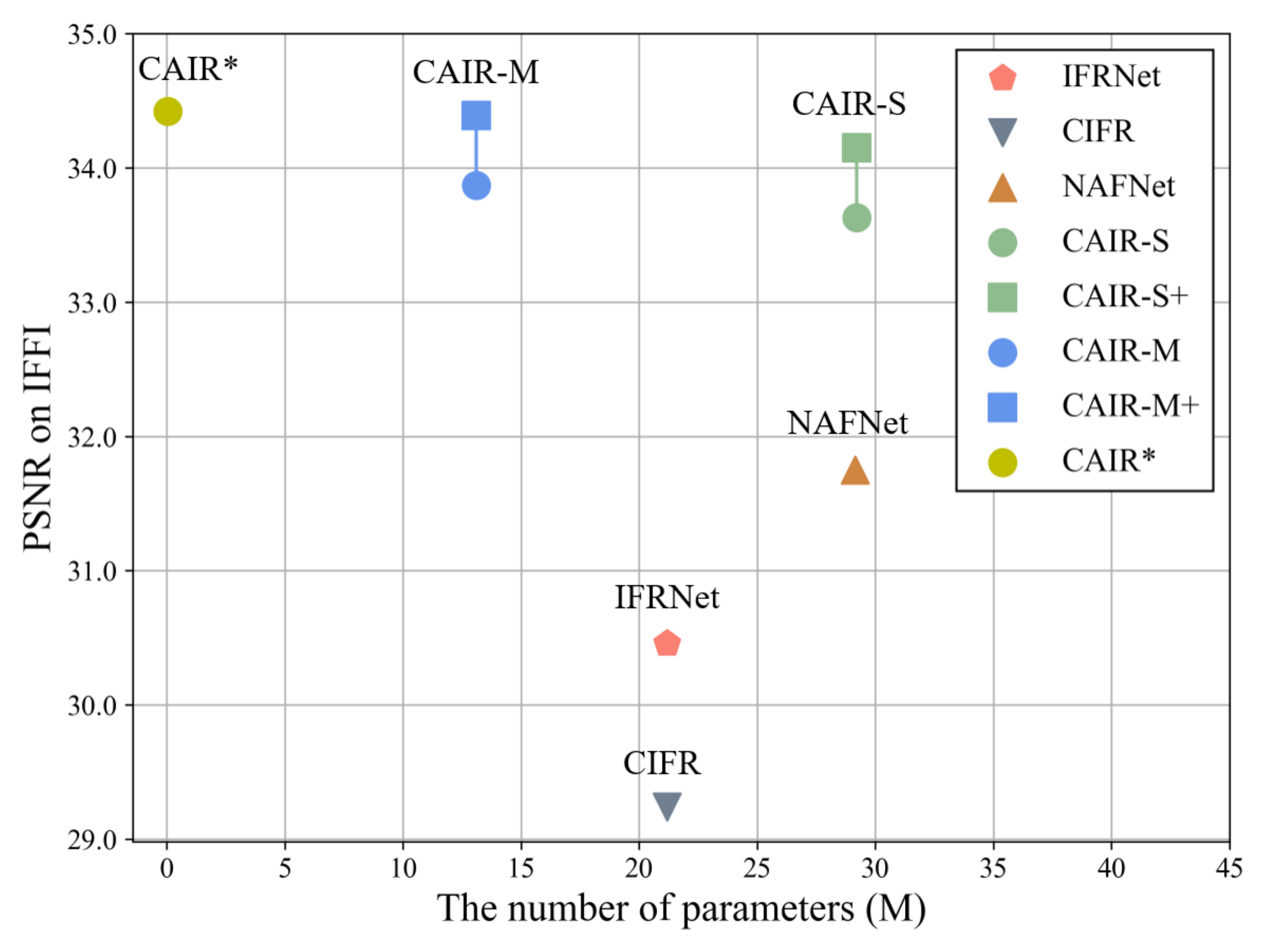}
        \subcaption{(b)}
    \end{subfigure}
    \caption{Comparison of the proposed CAIR and other methods. (a) PSNR vs. runtime (ms) trade-off, (b) PSNR vs. the number of parameters (M) on IFFI test sets tested on $4\times$ NVIDIA Tesla V100.}
    \label{fig:tradeoff}
\end{figure}

Fig.~\ref{fig:tradeoff} shows the PSNR vs. inference time tradeoffs and PSNR vs. the number of parameters. CAIR is much faster and has fewer parameters than the previous Instagram filter removal methods~\cite{kinli2021instagram,kinli2022patch} while attaining more accurate results. The PSNR and SSIM of IFRNet and CIFR were obtained from the results in~\cite{kinli2021instagram,kinli2022patch}. The runtime of all methods was measured on the same machine ($4\times$ NVIDIA Tesla V100). Although the inference time of NAFNet and the number of parameters are not significantly different from CAIR, it should be noted that CAIR exceeds PSNR and SSIM. The inference time and the number of parameters on CAIR* were measured only for inferencing with the ensemble network. CAIR* would take more time and computational complexity than the values in Table~\ref{tab:res}, since ensemble learning is followed by inferencing images with two models; CAIR-S and CAIR-M).

\subsubsection{CAIR for AIM 2022 challenge}
Table~\ref{tab:challenge} shows the results of the AIM 2022 challenge on Instagram filter removal (IFR). Compared with the two baselines, IFRNet and CIFR, our method achieved significant improvement in all metrics. Our team placed \nth{5} in the AIM 2022 challenge on IFR~\cite{kinli2022aim}. However, in terms of inference speed, our proposed method was more than 10$\times$ the fastest in the top 5 and ranked second among all participants, as shown in Table~\ref{tab:challenge}. The model structure and training strategy are the same as described above, except that the ensemble learning strategy is slightly different. For ensemble learning, images were trained on three models, two CAIR-Ss of width 32 and 64 and one CAIR-M of width 32. Furthermore, the outputs from the three models were concatenated and fed into the ensemble network. The challenge ensemble results in Table~\ref{tab:challenge} are worse than the CAIR* in Table~\ref{tab:res}. This result indicates that learning results can vary depending on how models are combined for ensemble learning. Therefore, when using the ensemble strategy in practice, it should be used carefully while checking the results.

\setlength{\tabcolsep}{10pt}
\begin{table}[htbp]
\centering
\caption{Results of AIM 2022 Instagram filter removal challenge~\cite{kinli2022aim}.}
\begin{tabular}{lccccc}
\toprule
\text{Team name} & \text{PSNR} & \text{SSIM} & \text{Runtime (s)} & \text{CPU/GPU} \\ \midrule
Fivewin & \bftab{34.70} & \bftab{0.97} & 0.91 & GPU \\
CASIA LCVG & 34.48 & 0.96 & 0.43 & GPU \\
MiAlgo & 34.47 & \bftab{0.97} & 0.40 & GPU \\
Strawberry & 34.39 & 0.96 & 10.47 & GPU \\
SYU-HnVLab(ours) & 33.41 & 0.95 & 0.04 & GPU\\
XDER & 32.19 & 0.95 & 0.05 & GPU \\
CVRG & 31.78 & 0.95 & 0.06 & GPU \\
CVML & 30.93 & 0.94 & \bftab{0.02} & GPU \\
Couger AI & 30.83 & 0.94 & 10.43 & CPU \\
IFRNet \cite{kinli2021instagram} & 30.46 & - & 0.60 & GPU \\
(Baseline1) &&&& \\
CIFR \cite{kinli2022patch} & 30.02 & - & 0.62 & GPU \\ 
(Baseline2) &&&& \\\bottomrule

\end{tabular}%
\label{tab:challenge}
\end{table}

%% file: article/conclusion.tex
This paper introduced CAIR, an end-to-end lightweight filter removal network suitable for Instagram filters. CAIR was based on the simple baseline for image restoration NAFNet. The proposed color attention module, inspired by the color correction branch of CycleISP, can extract color-informative features from multi-scale input images so that filters can be effectively removed from filtered images. We achieved fast (45.14 ms inference time), lightweight (15.23 GMACs), and accurate (34.15 dB PSNR) performance with CAIR-M. The results were improved further by the self-ensemble and the ensemble learning strategy. Experimental results on the IFFI dataset verify that the proposed method is efficient and capable of achieving high speed and accuracy among the compared methods. Especially, CAIR restores color information during filter removal much better than the other methods. It is because the color attention modules are trained to extract color-attentive features with scale-invariant features. By extending the scope of the dataset for all available Instagram filters, this method could be employed for pre-processing the social media images before feeding them into a vision framework to enhance its performance.

%% file: article/ack.tex
This research was supported by Basic Science Research Program through the National Research Foundation (NRF) funded by the Ministry of Science and ICT (grant no. NRF-2021R1F1A1059493) and Electronics and Telecommunications Research Institute (ETRI) grant funded by the Korea government (grant no. 22ZB1140, Development of Creative Technology for ICT).